\documentclass[10pt,twocolumn,letterpaper]{article}

\usepackage[pagenumbers]{cvpr} 

\definecolor{cvprblue}{rgb}{0.21,0.49,0.74}
\usepackage[pagebackref,breaklinks,colorlinks,allcolors=cvprblue]{hyperref}
\usepackage{algorithm}
\usepackage{algorithmic}
\usepackage{amsmath}
\usepackage{booktabs}
\usepackage{array}
\usepackage{enumitem}
\usepackage{makecell}
\usepackage{ragged2e}
\usepackage{multirow}
\usepackage{newfloat}
\usepackage{listings}
\usepackage{subcaption}
\usepackage{amssymb}
\usepackage{pifont}
\usepackage[accsupp]{axessibility}

\def\paperID{10378} 
\def\confName{CVPR}
\def\confYear{2025}

\title{DocLayLLM: An Efficient Multi-modal Extension of Large Language Models for Text-rich Document Understanding}

\author{
    Wenhui Liao\textsuperscript{\rm 1}$^{\dagger}$,
    Jiapeng Wang\textsuperscript{\rm 1,2}$^{\dagger}$,
    Hongliang Li\textsuperscript{\rm 1},
    Chengyu Wang\textsuperscript{\rm 2}$^{*}$,
    Jun Huang\textsuperscript{\rm 2},
    Lianwen Jin\textsuperscript{\rm 1,3}$^{*}$\\
    \textsuperscript{\rm 1}South China University of Technology, Guangzhou, China\\
    \textsuperscript{\rm 2}Alibaba Cloud Computing, Hangzhou, China\\
    \textsuperscript{\rm 3}SCUT-Zhuhai Institute of Modern Industrial Innovation, Zhuhai, China\\
    {\tt\small \{eelwh, eejpwang, eehongliangli\}@mail.scut.edu.cn, eelwjin@scut.edu.cn}\\
    {\tt\small{\{chengyu.wcy, huangjun.hj\}@alibaba-inc.com}}
}

\begin{document}
\maketitle
\begin{abstract}
Text-rich document understanding (TDU) requires comprehensive analysis of documents containing substantial textual content and complex layouts. While Multimodal Large Language Models (MLLMs) have achieved fast progress in this domain, existing approaches either demand significant computational resources or struggle with effective multi-modal integration.
In this paper, we introduce \textbf{DocLayLLM}, an efficient multi-modal extension of LLMs specifically designed for TDU. By lightly integrating visual patch tokens and 2D positional tokens into LLMs' input and encoding the document content using the LLMs themselves, we fully take advantage of the document comprehension capability of LLMs and enhance their perception of OCR information. We have also deeply considered the role of chain-of-thought (CoT) and innovatively proposed the techniques of \textit{CoT Pre-training} and \textit{CoT Annealing}. Our DocLayLLM can achieve remarkable performances with lightweight training settings, showcasing its efficiency and effectiveness. Experimental results demonstrate that our DocLayLLM outperforms existing OCR-dependent methods and OCR-free competitors. Code and model are available at \href{https://github.com/whlscut/DocLayLLM}{https://github.com/whlscut/DocLayLLM}.
\end{abstract}

\renewcommand{\thefootnote}{\fnsymbol{footnote}}
\footnotetext[2]{Equal contribution.}
\footnotetext[1]{Co-corresponding authors.}
\renewcommand{\thefootnote}{\arabic{footnote}}    
\section{Introduction}
\label{sec:intro}

Text-rich document understanding (TDU) focuses on analyzing documents with extensive text and complex layouts, requiring a combined understanding of text, layout, and images. Early research~\citep{xu2020layoutlm, li2021structurallm, li2021selfdoc, Powalski2021tilt, lukasz2021lambert, xu2021layoutlmv2, Appalaraju_2021_ICCV, li2021structext, wang2022lilt, wang2025liltv2, li2022docrel, peng2022ernie, hong2022bros, huang2022layoutlmv3, tu2023layoutmask, zhang2023reading, Luo_2023_CVPR, li2023graphlayoutlm, appalaraju2024docformerv2, li2024hypergraph} often used BERT-like~\citep{jacob2019bert} Transformer-based document encoders (DE), designing various pre-training tasks for large-scale unlabeled document corpora and fine-tuning their models on specific downstream tasks like key information extraction (KIE) and visual question answering (VQA)~\citep{lin2023visual, liu2023frontiers}. However, these pre-training tasks often diverge significantly from downstream tasks, requiring the model to acquire substantial task-specific knowledge during fine-tuning. This may result in limited generalization capability across different tasks and datasets.

With the rapid development of large language models (LLMs), their impressive versatility and generalization capabilities have captured widespread interest~\citep{yan2023development}. This has led many researchers~\citep{zhang2023llavar, ye2023ureader, tang2024textsquare, hu2024mplugdocowl1_5, li2024monkey} to overcome the challenges of TDU with LLMs. Initially, efforts focused on employing multi-modal large language models (MLLMs) for OCR-free TDU, where document images are fed into MLLMs, directly generating responses to comprehension tasks. These MLLMs, which can process images and text inputs, are among the most recent advancements in this domain. To date, several general MLLMs have been developed to manage diverse scenarios, and researchers often build their TDU models based on these general models.
However, these TDU models typically require high-resolution image inputs to recognize dense text in documents and a large volume of training data to effectively develop this capability, both of which bring significant computational demands. Furthermore, within the extensive collections of image and instruction data used to train these models, the training data for downstream evaluation tasks was often included, which might potentially impact the assessment of their generalization to previously unseen datasets.

To overcome the constraints of OCR-free methods, scholars have pursued diverse methodologies for integrating OCR information into LLMs. Initial strategies~\citep{wang2023latinprompot, he2023icld3ie} involved inputting coordinates of OCR bounding boxes as numeric text. However, this textual representation of numbers often resulted in excessively long inputs that could decrease inference speeds and potentially degrade model performance. In response, many researchers~\cite{kim2023cream, fujitake2024layoutllm, luo2024layoutllm} have adopted an auxiliary document encoder like LayoutLMv3~\citep{huang2022layoutlmv3} to process textual, visual, and layout features from documents and integrated into LLMs. Although these methods avoid high-resolution image inputs, the textual content of documents must also be encoded alongside the layout and image features within the DE due to its structural limitations. However, the document comprehension capabilities of these DEs may not be on par with those of LLMs themselves, which might have led to significant fine-tuning efforts for aligning the feature spaces of DEs and LLMs to improve performances during the whole training phase, demanding substantial computational resources. 

To address the limitations of OCR-dependent TDU methods, we introduce \textbf{DocLayLLM}, an efficient multimodal extension of large language models for text-rich document understanding. Our method diverges from existing approaches that utilize an extra DE by integrating the document's image patch tokens and 2D positional tokens with textual content into a natural language expression, and then encoding them through a unified LLM. This innovative integration not only taps into the LLM’s inherent text comprehension abilities but also allows it to perceive document layouts. Besides, by incorporating patch tokens, the model supports image-related pre-training tasks and further gains a rudimentary understanding of document structures.

To advance model performance, we have fully integrated the concept of the chain-of-thought (CoT)~\citep{tom2023gpt3} into each stage of our training process. Firstly, we developed a novel \textit{CoT Pre-training} strategy. CoT is engineered to preserve thematic coherence and logical progression during model reasoning, further enhancing the inference ability of LLM, which has proven effectiveness in various subsequent studies~\citep{jason2022cotprompting, he2023icld3ie, luo2024layoutllm}.
Recently, the use of Long CoT reasoning to enhance the inference capability of models has attracted significant attention~\citep{gpt4o2024openai, qwq2024qwen, deepseekr12025deepseek}, and DeepSeek-R1~\citep{deepseekr12025deepseek} further emphasized the importance of utilizing CoT data for a cold start before the reinforcement learning stage. Our proposed CoT Pre-training offers a method for rapidly generating large-scale CoT data from non-QA data, thereby improving the model's reasoning ability.
By employing CoT in the data generation process and subsequently pre-training the model with the generated QA-formatted data, we have aligned the added visual and layout features with LLM's inherent textual features while improving the model’s comprehension ability.
Furthermore, we introduced a \textit{CoT Annealing} technique. While previous research~\citep{li2024datacomplm, llama32024llama3_1} has validated the effectiveness of annealing strategies that gradually increase the proportion of high-quality data towards the end of training, the impact of CoT on data quality had not been explored. To address this, we first propose an annealing strategy that considers data quality from a CoT perspective, further improving model performance.

Thanks to the above design, utilizing only around 3M pre-training data and 300K supervised fine-tuning (SFT) data with the efficient tuning technology of LoRA~\citep{hu2022lora}, our DocLayLLM finishes the whole training process in 36 A100 days,
therefore resulting an efficient extension of LLM. Even with such low resources, our DocLayLLM surpasses existing OCR-dependent TDU methods. Furthermore, it also excels in specific tasks without fine-tuning on downstream tasks and outperforms state-of-the-art (SOTA) OCR-free methods when trained with training sets of downstream tasks as other methods, further showcasing its effectiveness. Our contributions are:

\begin{itemize}
    \item 
    We propose an efficient multi-modal extension of LLMs, which has augmented LLM in understanding text-rich documents and greatly reduced training resources in need.
    \item 
    We introduce \textit{CoT Pre-training}, a novel approach for generating CoT data that helps enhance the comprehension capabilities of models.
    \item 
    We revisit data annealing from a novel perspective and introduce the \textit{CoT Annealing} strategy, which further enhances the model's efficiency in data utilization and boosts its ultimate performance.
    \item 
    Our method outperforms existing OCR-dependent TDU solutions and surpasses current OCR-free methods under comparable supervised fine-tuning conditions.
\end{itemize}

\section{Related Works}
\label{sec:related works}

\begin{figure*}[t]
\centering
\includegraphics[width=0.875\textwidth]{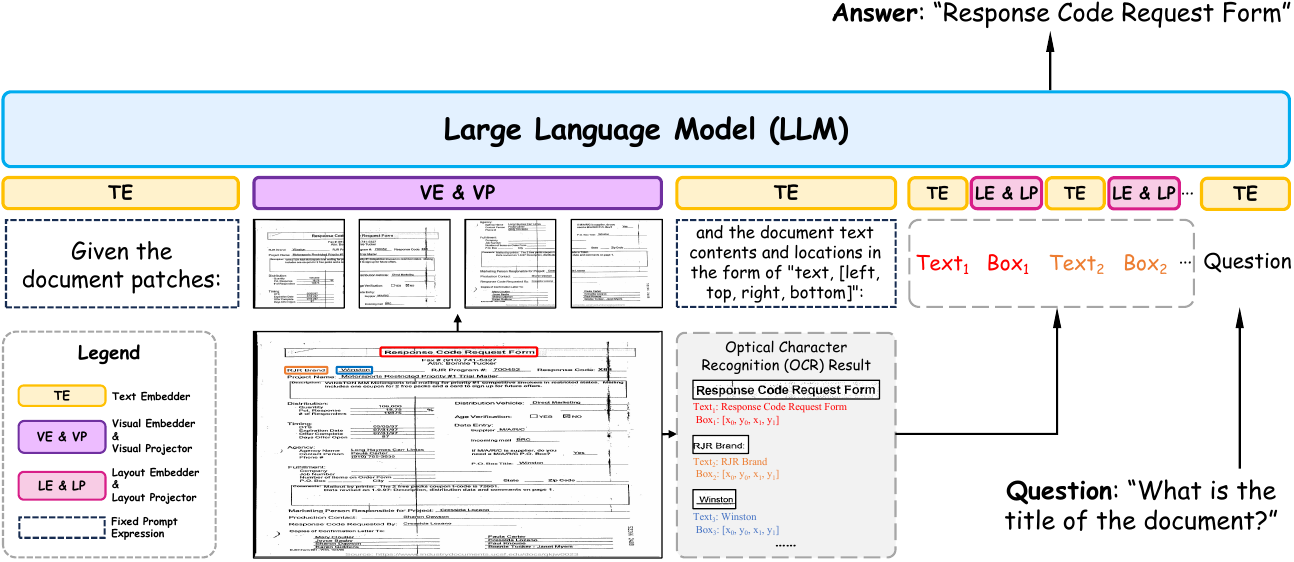}
\caption{The overall architecture of DocLayLLM.}
\label{overall_architecture}
\end{figure*}

\textbf{OCR-free MLLMs for Text-Rich Documents Understanding.}\ \ 
To date, a number of studies have attempted to directly answer the instruction for TDU tasks from document images, thereby obviating the need for extra OCR processing and resulting in an end-to-end paradigm. Such models are referred to as OCR-free methods. Early examples like Donut~\citep{kim2022donut}, UDOP~\citep{tang2023udop}, and OmniParser~\citep{wang2024omniparser} have explored end-to-end document understanding solutions without relying on LLMs or MLLMs. However, with the rapid development of MLLMs, many new studies have begun to focus on enhancing the capabilities of general MLLMs in document understanding. 
Some of them attempt to boost the document understanding ability of general MLLMs by tuning them with synthesizing TDU data generated by proprietary LLMs, including LLaVAR~\citep{zhang2023llavar} and TextSquare~\citep{tang2024textsquare},
while mPLUG-DocOwl 1.5~\citep{hu2024mplugdocowl1_5} enhances the document understanding capability by integrating existing document comprehension data and converting it into vision-related QA-formatted data. Moreover, researchers have also explored image encoding methods to read the text more clearly. For instance, UReader~\cite{ye2023ureader} and Monkey~\cite{li2024monkey} empower MLLMs to handle high-resolution images by splitting them into sub-images, encoding each separately, and then concatenating them.
Some general MLLMs, such as LLaVA-Next~\citep{li2024llavanext}, LLaVA-OneVision~\citep{li2025llavaonevision}, and Idefics3~\citep{laurencon2024building}, also adopt this approach to address the challenge of TDU.
Although these above methods have achieved notable success in TDU tasks, they typically require high-resolution image inputs to identify text within images, significantly increasing computational demands. Besides, to equip general MLLMs with the ability to recognize the rich text within a document, a substantial amount of training data is required, including training sets of the benchmarks tested, which resulted in a high demand for training resources and might affect the evaluation of their zero-shot inference ability.

\textbf{OCR-dependent MLLMs for Text-Rich Documents Understanding.}\ \ 
To address the challenges of OCR-free MLLMs, some researchers have attempted to integrate OCR information from text-rich documents into LLMs to avoid high-resolution image inputs. ICL-D3IE~\citep{he2023icld3ie} feeds coordinates of OCR boxes into the LLM using a textual expression of \textit{[left, top, right, bottom]}, while LATIN-Prompt~\citep{wang2023latinprompot} simulates the document layout by inserting spaces and newline characters between OCR-recognized text segments, thus providing the LLM with a document input containing a rudimentary layout. Besides, LMDX~\citep{perot2024lmdx} also input the layout in a way similar to ICL-D3IE. However, these methods of inputting OCR information in textual form lead to overly lengthy inputs, thereby slowing inference and impacting model performance. Subsequent studies have explored encoding OCR information using additional encoders. Cream~\citep{kim2023cream} utilizes BLIP-2~\citep{li2023blip2} to encode document OCR positions, images, text, and additional object detection information, with the output of BLIP-2 fed into the LLM. InstructDr~\citep{tanaka2024instrucdoc} introduces an extra feature extractor to encode document OCR information, compressing it with learnable tokens before inputting it into the LLM. Additionally, ~\citet{fujitake2024layoutllm} and~\citet{luo2024layoutllm} adopt LayoutLMv3~\citep{huang2022layoutlmv3} to encode the OCR information of documents before inputting them into the LLM. Although these methods avoid the issue of overly lengthy text inputs, they must use the extra DE to encode all the OCR textual, positional information from the document due to the structural restriction of DE. However, the additional DE might not be comparable with LLMs themselves in terms of document comprehension capability. This might be the reason why they usually need extensive training of both the DE and the LLM to align the features encoded by the DE with those of the LLM, which consumes significant computational resources. 
DocLLM~\citep{wang2024docllm} proposes an extension where the hidden states derived from OCR bounding boxes interact with those of OCR text features encoded by LLMs within each attention block via cross-attention~\citep{dai2023disentangling, dai2025onedm}. However, they use an extra projection layer at each attention block to transform and inject layout information, which might excessively interfere with the original attention computation of LLM, thereby potentially affecting LLM’s generalization capability. Moreover, it does not adequately consider the role of image features in TDU. Overall, designing an efficient method to provide OCR information to LLMs remains a research question worthy of deep exploration.
\section{DocLayLLM}
\label{sec:doclayllm}

\begin{figure*}[t]
\centering
\includegraphics[width=0.92\textwidth]{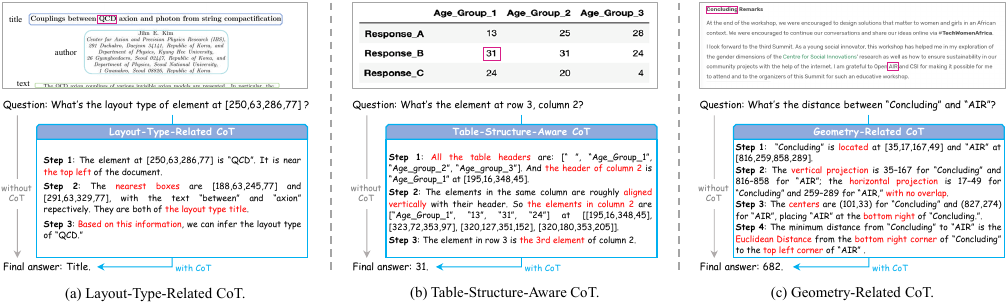}
\caption{Examples of CoT Pre-training across different tasks. Red text highlights key steps in the chain of thought for each task.}
\label{cot_pt}
\end{figure*}

DocLayLLM is an LLM-based methodology that encodes OCRed textual, visual, and positional information directly within an LLM, eliminating the need for additional DE. The training regime of DocLayLLM includes both pre-training and supervised fine-tuning phases, which are both instruction-tuning-based processes. To further enhance our model, we propose the integration of CoT into our pre-training tasks, denoted as \textit{CoT Pre-training}. Additionally, we introduce \textit{CoT Annealing}, which involves the application of annealing to the data utilized in the supervised fine-tuning phase, specifically from the perspective of CoT.

\subsection{Model Architecture}
The overall architecture of our DocLayLLM is illustrated in Figure~\ref{overall_architecture}. To process a document image, we first employ an OCR engine. This engine extracts the OCR results, which encompass the textual content, denoted as $Text_{1:N}$, and the associated bounding box, represented by $Box_{1:N}$. Here, $N$ indicates the total number of text segments detected by the OCR, and each $Box_i = [x_0, y_0, x_1, y_1]$ defines the bounding box for $Text_i$. The bounding box coordinates correspond to the \textit{left, top, right,} and \textit{bottom} boundaries of the text segment. After the OCR process, the document image is segmented into patches, denoted as $Patch_{1:M}$, where $M$ represents the total number of patches. Next, the OCR results and the document patches are integrated with the fixed part of the prompt and the specific $Question$ related to the document. This integration forms the final prompt input to the model, presented in a natural language expression $\mathcal{T}$: 
\textit{Given the document patches:} \{$Patch_{1:M}$\} \textit{and the document text contents and locations in the form of ``text, }[\textit{left, top, right, bottom}]\textit{''}: \{$(Text_i, Box_i)_{1:N}$\} \{$Question$\}

Subsequently, the textual components of the prompt, including the fixed expression, $Text_{1:N}$, and  $Question$, are processed using the LLM's native text embedder. This generates the embeddings for the text content as follows:
\begin{equation}
    Emb_{T}=TE(TextContent),
\end{equation}
where $TE$ is the native text embedder of LLM, and $TextContent$ consists of the fixed expression, $Text_{1:N}$, and $Question$.

As for $Patch_{1:M}$ and $Box_{1:N}$, they are first embedded with an extra embedder, then mapped to the LLM's feature space via a linear projector, yielding $Emb_{V}$ and $Emb_{L}$ as:
\begin{align}
    Emb_{V}&=VP(VE(Patch_{1:M})),\\
    Emb_{L}&=LP(LE(Box_{1:N})),
\end{align}
where $VE$ and $VP$ represent the visual embedder and visual projector, $LE$ and $LP$ denote the layout embedder and layout projector, respectively. It is worth noting that despite the introduction of $VE$, $VP$, $LE$, and $LP$, these are all simple embedding layers or linear layers, thereby minimizing additional computational burden.

Finally, LLM will comprehend the embeddings provided and utilize them to generate the $Answer$ to the $Question$:
\begin{equation}
\begin{split}
    Answer = LLM(\pi ((Emb_{T}, Emb_{V}, Emb_{L}), \mathcal{T})).
\end{split}
\end{equation}

Here $\pi$ represents organizing $Emb_{T}$, $Emb_{V}$, $Emb_{L}$ into a feature sequence according to $\mathcal{T}$.

With our design, we have preserved the original structure of the LLM, ensuring its generalization capability. At the same time, we have enabled the LLM to easily accept and process visual and layout features, making it better suited for understanding text-rich documents with various layouts. This results in a simple yet effective multi-modal extension.

\subsection{CoT Pre-training}
During the pre-training phase, we incorporate a wide range of tasks (as detailed in the appendix) to boost the LLM's understanding of the visual and layout features inspired by~\citet{luo2024layoutllm}. Since the datasets related to these tasks are not designed explicitly for instruction-tuning, their annotations are not in a QA format. Therefore, we manually create QA-formatted instructions based on their annotations for training. However, simply converting task descriptions into $Question$ and annotations into $Answer$ is highly inefficient regarding data utilization. To address this, we incorporate CoT during the conversion to instruction-tuning data for specific tasks, asking LLM to solve our problem step-by-step thus enhancing the model's comprehension abilities, as the examples illustrated in Fig~\ref{cot_pt}. Our designed CoT primarily falls into the following three categories:

\subsubsection{Layout-Type-Related CoT}
We conduct a layout-type-related CoT for the layout analysis task. To determine the layout type of a given area, it is beneficial first to locate its region within the document and then infer based on the nearby layout element. When giving an area bounded by a bounding box, our layout-type-related CoT can be formulated as following steps:
\begin{enumerate}
    \item Identify the text within the area and determine its approximate region within the document (e.g., top left, top right part of the document, etc).
    \item Locate the nearest bounding boxes and assess their layout type.
    \item Infer the layout type of the given area based on the above information.
\end{enumerate}
Through this CoT process, we enable the LLM to fully utilize the visual, layout, and textual features of the document, thereby improving their ability to understand the document's overall structure.

\subsubsection{Table-Structure-Aware CoT}
Table-structure-aware CoT is primarily employed in the table analysis task. We have integrated the XY-Cut algorithm~\cite{g1992xycut}—a widely-used technique for analyzing table structures. For instance, to locate an element at the $i$-th row and $j$-th column, the CoT process is outlined as follows:
\begin{enumerate}
    \item Identify all the table headers and their respective locations. The header of the $j$-th column corresponds to the $j$-th element in the headers list.
    \item Output all elements in the $j$-th column when told that the elements of the same column are roughly vertically aligned with their corresponding headers.
    \item Determine the final answer based on the $i$-th element of the contents in the $j$-th column.
\end{enumerate}
Applying the above CoT, we break down a complex question into simple steps using the basic XY-Cut thought. This approach improves the LLM's inference ability through this breakdown and enhances data utilization by effectively leveraging the overall structure of the table.

\subsubsection{Geometry-Related CoT}
This is used in geometric analysis to identify geometric relationships between two text segments.
It assists LLM to inference with geometric relations.
Specifically, when giving the text content of two lines:
\begin{enumerate}
    \item Recover the bounding box of the given text segments.
    \item Analyze their vertical and horizontal projections to determine if they overlap through interval analysis.
    \item Calculate the center point of each segment and determine their relative location by comparing their center points.
    \item Calculate the minimum Euclidean distance based on the previous analysis. If they overlap, the distance is 0; if they overlap in vertical/horizontal projections, the minimum distance is the smallest interval in the horizontal/vertical projections; otherwise, it is the minimum Euclidean distance between their nearest corners.
\end{enumerate}
This design advances the LLM's arithmetic reasoning capabilities and helps it more accurately grasp the spatial relationships between different boxes, and also contributes to layout-type-aware CoT to find the nearest bounding boxes.

\subsection{CoT Annealing}
Data annealing involves gradually increasing the proportion of high-quality data towards the end of supervised fine-tuning to improve data utilization efficiency and augment the model's reasoning abilities. This technique has been proven to be effective and is widely adopted in models like MiniCPM~\citep{hu2024minicpm}, DataComp-LM~\citep{li2024datacomplm}, and Llama-3.1~\citep{llama32024llama3_1}.

During the SFT stage, we employ data with a layout-aware CoT~\cite{luo2024layoutllm} that helps focus on potential regions of questions and answers within documents, thereby better leveraging the layout features learned during pre-training. It also allows the model to strengthen its reasoning capacity through step-by-step thinking. However, in the latter stages of SFT, our goal for LLM is to output precise and direct answers to questions. At this point, using CoT data may introduce unnecessary noise with a long response, and the overly uniform layout-aware CoT data might partially compromise the LLM's generalization abilities.

To balance the strengths and weaknesses of such data, we propose CoT annealing, which reconsiders this data annealing issue from a CoT perspective. Specifically, we generate direct-answer data without CoT from the CoT data, treating it as high-quality data. 
During the SFT process, we gradually increase the proportion of this direct-answer data, thereby enhancing the LLM's ability to respond with a clear and direct answer in the end. Initially, we only utilize the data with CoT, gradually adjusting the ratio of CoT to non-CoT data as the SFT process progresses. Ultimately, we conclude the training using only the data without CoT.

\section{Experiments}
\label{sec:experiments}

\begin{table*}[t]
\centering
\resizebox{.81\linewidth}{!}{
    \begin{tabular}{l|c|ccc|cccc}
    \toprule[2pt]
     \multirow{2}{*}{\textbf{Method}} & \multirow{2}{*}{\makecell[c]{\textbf{Processed Data }\\ \textbf{at Pre-training}}} & \multicolumn{3}{c|}{\textbf{Document-oriented VQA}} & \multicolumn{4}{c}{\textbf{KIE}} \\
          &       & DocVQA & VisualMRC & \textbf{Avg.} & FUNSD$^{\dagger}$ & CORD$^{\dagger}$  & SROIE$^{\dagger}$ & \textbf{Avg.} \\
    \midrule
    \textbf{Text} &       &       &       &       &       &       &       &  \\
    Llama2-7B-Chat~\citep{touvron2023llama2} & -     & 20.50  & 9.90  & 15.20  & 15.10  & 20.00  & 35.6  & 23.57  \\
    Llama3-8B-Instruct~\citep{llama32024llama3_1} & -     & 51.79  & 47.77  & 49.78  & 68.56  & 52.31  & 61.24  & 60.70  \\
    \midrule
    \textbf{Text + Box} &       &       &       &       &       &       &       &  \\
    Llama2-7B-Chat$_{coor}$~\citep{he2023icld3ie} & -     & 12.30  & 12.20  & 12.25  & 11.90  & 6.40  & 39.40  & 19.23  \\
    Llama3-8B-Instruct$_{coor}$~\citep{he2023icld3ie} & -     & 49.13  & 41.75  & 45.44  & 74.00  & 62.20  & 63.15  & 66.45  \\
    LayTextLLM$_{zero}$~\citep{lu2024laytextllm} & 10.0M (323\%) & 65.50  & 37.40  & 51.45  & 72.00  & 45.50  & 82.00  & 66.50  \\
    \midrule
    \textbf{Text + Box + Patch} &       &       &       &       &       &       &       &  \\
    LayoutLLM$_{Llama2}$~\citep{luo2024layoutllm} & 5.7M (184\%) & \underline{74.25}  & \underline{55.73}  & \underline{64.99}  & 78.65  & 62.21  & 70.97  & 70.61  \\
    \rowcolor[rgb]{ .741,  .843,  .933} $DocLayLLM_{Llama2}$ \textbf{(Ours)} & \textbf{3.1M (100\%)} & 72.83  & 55.04  & 63.94  & \underline{78.67}  & \underline{70.81}  & \underline{83.27}  & \underline{77.58}  \\
    \rowcolor[rgb]{ .741,  .843,  .933} $DocLayLLM_{Llama3}$ \textbf{(Ours)}  & \textbf{3.1M (100\%)} & \textbf{78.36 } & \textbf{58.55 } & \textbf{68.46 } & \textbf{84.11 } & \textbf{71.34 } & \textbf{84.36 } & \textbf{79.94 } \\
    \bottomrule[2pt]
    \end{tabular}
}
\caption{Comparison with OCR-dependent methods using metrics from LayoutLLM~\cite{luo2024layoutllm}.
\textbf{Processed Data at Pre-training} represents the total data volume processed at the pre-training stage, equivalent to dataset size multiplied by epoch count, with values in parentheses representing size ratios compared to DocLayLLM.
\textbf{Avg.} represents the average performance across different benchmarks, and $^{\dagger}$ indicates the cleaned test set used in LayoutLLM.
The suffix $_{layout}$ denotes using method from~\citet{he2023icld3ie}, which provides LLM with layout features via OCR box coordinates in textual form.
\textbf{Bold} indicates the best performance, and \underline{underline} indicates the second-best one.}
\label{com_with_layoutllm}
\end{table*}

\begin{table*}[t]
\centering
\resizebox{.81\linewidth}{!}{
    \begin{tabular}{l|c|ccc|cccc}
    \toprule[2pt]
    \multirow{2}{*}{\textbf{Method}} & \multirow{2}{*}{\makecell[c]{\textbf{Processed Data }\\ \textbf{at Pre-training}}} & \multicolumn{3}{c|}{\textbf{Document-oriented VQA}} & \multicolumn{4}{c}{\textbf{KIE}} \\
          &       & DocVQA & VisualMRC & \textbf{Avg.} & FUNSD & CORD  & SROIE & \textbf{Avg.} \\
    \midrule
    \textbf{Text} &       &       &       &       &       &       &       &  \\
    Llama2-7B-Chat~\citep{touvron2023llama2} &   -   & 20.50  & 6.30  & 13.40  & 23.40  & 51.80  & 58.60  & 44.60  \\
    Llama3-8B-Instruct~\citep{llama32024llama3_1} &  -    & 51.79  & 229.74  & 140.77  & 66.67  & 74.71  & 82.51  & 74.63  \\
    \midrule
    \textbf{Text + Box} &       &       &       &       &       &       &       &  \\
    Llama2-7B-Chat$_{coor}$~\citep{he2023icld3ie} &   -   & 12.30  & 28.00  & 20.15  & 14.40  & 38.10  & 50.60  & 34.37  \\
    Llama3-8B-Instruct$_{coor}$~\citep{he2023icld3ie} &   -   & 49.13  & 211.69  & 130.41  & 74.71  & 75.33  & 75.93  & 75.32  \\
    LMDX~\citep{perot2024lmdx} & - & -  & - & -  & -  & 66.95  & -  & -  \\
    DocLLM~\citep{wang2024docllm} & 16.8M (542\%) & 69.50$^*$  & 264.10$^*$ & 166.80$^*$  & 51.80$^*$  & 67.60$^*$  & \textbf{91.90$^*$}  & 70.43$^*$  \\
    LayTextLLM$_{zero}$~\citep{lu2024laytextllm} & 10.0M (323\%) & 65.50  & 200.20  & 132.85  & 47.20  & 77.20  & 83.80  & 69.40  \\
    \midrule
    \textbf{Text + Box + Patch} &       &       &       &       &       &       &       &  \\
    \rowcolor[rgb]{ .741,  .843,  .933} $DocLayLLM_{Llama2}$ \textbf{(Ours)} & \textbf{3.1M (100\%)} & \underline{72.83}  & \underline{310.60}  & \underline{191.72}  & \underline{80.74}  & \underline{79.37}  & 84.37  & \underline{81.49}  \\
    \rowcolor[rgb]{ .741,  .843,  .933} $DocLayLLM_{Llama3}$ \textbf{(Ours)} & \textbf{3.1M (100\%)} & \textbf{78.35 } & \textbf{357.04 } & \textbf{217.70 } & \textbf{86.47 } & \textbf{83.66 } & \underline{86.08 } & \textbf{85.40 } \\
    \bottomrule[2pt]
    \end{tabular}
}
\caption{Comparison with OCR-dependent methods using metrics from DocLLM~\cite{wang2024docllm}. 
$^{*}$ indicates that the model uses the training set of the benchmark during training.}
\label{com_with_docllm}
\end{table*}

\subsection{Datasets}
During the pre-training stage, we conduct our pre-training tasks on several datasets, including RVL-CDIP~\citep{harley2015rvlcdip}, PubLayNet~\citep{zhong20219publaynet}, PubTabNet~\citep{zhong2019pubtabnet}, DocLayNet~\citep{pfitzmann2022doclaynet}, DocBank~\citep{li2020docbank}, DocILE~\citep{dsimsa2023docile}, and the Document Dense Description data from LayoutLLM~\citep{luo2024layoutllm}. This results in a total of 3.1 million pairs of instruction tuning data, which is significantly less than the volume used in previous OCR-dependent TDU methods. For the supervised fine-tuning stage, we apply CoT annealing on the layout-aware document QA data utilized in LayoutLLM~\citep{luo2024layoutllm}.

We evaluate our DocLayLLM on widely used text-rich document benchmarks, including FUNSD~\citep{jaume2019funsd}, CORD~\citep{park2019cord}, SROIE~\citep{huang2019sroie}, POIE~\citep{kuang2023poie}, DocVQA~\citep{mathew2021docvqa}, InfoVQA~\citep{mathew2021infovqa}, VisualMRC~\citep{tanaka2021visualmrc}, DeepForm~\citep{svetlichnaya2020deepform}, KLC~\citep{Stanislawek2021klc}, WTQ~\citep{pasupat2015wtq}, and TabFact~\citep{chen2020tabfact}. For a fair comparison with other approaches, we use the default OCR data from the official datasets otherwise commercial OCR engines~\footnote{\href{https://www.textin.com/}{https://www.textin.com/}} if not provided, and follow their evaluation pipeline and metrics.

\subsection{Implement Details}
Our method is based on Llama3-8B-Instruct~\citep{llama32024llama3_1}, with a Llama2-7B-Chat~\citep{touvron2023llama2} version for better comparison with other methods. They are denoted as DocLayLLM$_{Llama3}$ and DocLayLLM$_{Llama2}$ respectively. We initialize $VE$ and $LE$ with weights from LayoutLMv3~\citep{huang2022layoutlmv3} to leverage document knowledge acquired during its pre-training, while $VP$ and $LP$ are randomly initialized. Besides, the image resolution is fixed at 224*224, resulting in 196 visual patch tokens, to capture an overview of document layout while bringing less computation burden. During pre-training and fine-tuning, we use LoRA~\cite{hu2022lora} with a rank of 64 on LLM and keep $VE$, $VP$, $LE$, and $LP$ trainable, leading to few tuning parameters. We train our model on 8 NVIDIA Tesla A100-80G GPUs, taking about 30 A100 days for pre-training and 6 A100 days for supervised fine-tuning. Additional details can be found in the appendix.

\begin{table*}[t]
\centering
\resizebox{0.76\linewidth}{!}{
\setlength{\tabcolsep}{3.0mm}{
    \begin{tabular}{l|ccc|cccc}
    \toprule[2pt]
          \multirow{2}{*}{\textbf{Method}} & \multicolumn{3}{c|}{\textbf{Document-oriented VQA}} & \multicolumn{4}{c}{\textbf{KIE}} \\
          & DocVQA & InfoVQA & \textbf{Avg.}  & FUNSD & SROIE & POIE  & \textbf{Avg.} \\
    \midrule
    LLaVAR~\citep{zhang2023llavar} & 12.30  & 16.50  & 14.40  & 0.50  & 5.20  & 5.90  & 3.87  \\
    UniDoc~\citep{feng2023unidoc} & 7.70  & 14.70  & 11.20  & 1.00  & 2.90  & 5.10  & 3.00  \\
    DocPedia~\citep{feng2023docpedia} & 47.10$^{*}$  & 15.20$^{*}$  & 31.15$^{*}$  & 29.90  & 21.40  & 39.90  & 30.40  \\
    Monkey~\citep{li2024monkey} & 50.10$^{*}$  & 25.80$^{*}$  & 37.95$^{*}$  & 24.10  & 41.90  & 19.90  & 28.63  \\
    TextMonkey~\citep{liu2024textmonkey} & \underline{66.70$^{*}$}  & 28.60$^{*}$  & 47.65$^{*}$  & 42.90  & 46.20  & 32.00  & 40.37  \\
    DocOwl 1.5~\citep{hu2024mplugdocowl1_5} & -     & -     & -    & -   & 48.30  & 51.80     & - \\
    TextSquare~\citep{tang2024textsquare} & -     & -     & -     & -   & \underline{53.20}  & \underline{71.80}    & - \\
    DocKylin~\citep{zhang2024dockylin} & 65.10$^{*}$  & \underline{34.80$^{*}$}  & \underline{49.95$^{*}$}  & \underline{58.70}  & 25.50  & 49.50  & \underline{44.57}  \\
    \midrule
    \rowcolor[rgb]{ .741,  .843,  .933} $DocLayLLM_{Llama3}$ \textbf{(Ours)} & \textbf{77.79} & \textbf{42.02} & \textbf{59.91} & \textbf{80.28} & \textbf{76.59} & \textbf{75.13} & \textbf{77.33} \\
    \bottomrule[2pt]
    \end{tabular}
}}
\caption{Comparison with OCR-free methods using the precision metric from TextMonkey~\cite{liu2024textmonkey}. $^{*}$ indicates that the model uses the training set of
the benchmark during training.
}
\label{com_with_textmonkey}
\end{table*}

\begin{table*}[t]
\centering
\resizebox{0.89\linewidth}{!}{
    \begin{tabular}{l|cccc|ccc|ccc}
    \toprule[2pt]
          \multirow{2}{*}{\textbf{Method}} & \multicolumn{4}{c|}{\textbf{Document-oriented VQA}} & \multicolumn{3}{c|}{\textbf{KIE}} & \multicolumn{3}{c}{\textbf{Table Understanding}} \\
          & DocVQA & InfoVQA & VisualMRC & \textbf{Avg.} & DeepForm & KLC   & \textbf{Avg.} & WTQ   & TabFact & \textbf{Avg.} \\
    \midrule
    DocPedia~\citep{feng2023docpedia} & 47.10  & 15.20  & -     & -     & -     & -     & -     & -     & -     & - \\
    DocOwl~\citep{ye2023docowl} & 62.20  & 38.20  & 188.80  & 96.40  & 42.60  & 30.30  & 36.45  & 26.90  & 60.20  & 43.55  \\
    UReader~\citep{ye2023ureader} & 65.40  & 42.20  & 221.70  & 109.77  & 49.50  & 32.80  & 41.15  & 29.40  & 67.60  & 48.50  \\
    Monkey~\citep{li2024monkey} & 66.50  & 36.10  & -     & -     & 40.60 & 32.80  & 36.70     & 25.30  & -     & - \\
    TextMonkey~\citep{liu2024textmonkey} & 73.00  & -     & -     & -     & 59.70  & 37.80  & 48.75     & 31.90  & -     & - \\
    DocKylin~\citep{zhang2024dockylin} & 77.30  & 46.60  & \underline{319.90}  & \underline{147.93}  & 64.20  & -     & -     & 32.40  & -     & - \\
    CogAgent~\citep{hong2024cogageng} & 81.60  & 44.50  & -     & -     & -     & -     & -     & -     & -     & - \\
    DocOwl 1.5~\citep{hu2024mplugdocowl1_5} & 82.20  & 50.70  & 246.40  & 126.43  & \underline{68.80}  & \underline{38.70}  & \underline{53.75}  & 40.60  & 80.20  & 60.40  \\
    TextSquare~\citep{tang2024textsquare} & \underline{84.30}  & \underline{51.50}  & -     &   -   & -     & -     & -     & \underline{49.70}  & \textbf{84.20} & \underline{66.95}  \\
    \midrule
    \rowcolor[rgb]{ .741,  .843,  .933} $DocLayLLM_{Llama3}^{\ddagger}$ \textbf{(Ours)} & \textbf{86.52} & \textbf{58.36} & \textbf{327.91} & \textbf{157.60} & \textbf{77.07} & \textbf{40.73} & \textbf{58.90} & \textbf{58.62} & \underline{83.36}  & \textbf{70.99} \\
    \bottomrule[2pt]
    \end{tabular}
}
\caption{Comparison with OCR-free methods using the evaluation framework and metrics from DocOwl 1.5~\cite{hu2024mplugdocowl1_5}. All methods, including our $DocLayLLM_{Llama3}^{\ddagger}$ variant, are exposed to the train set corresponding to the evaluation set during training.
}
\label{com_with_docowl}
\end{table*}

\subsection{Comparisons with SOTA OCR-dependent methods}
We have compared our DocLayLLM with existing SOTA OCR-dependent methods. In order to exhibit the zero-shot inferring ability precisely, we chose their performances under the zero-shot scenario unless they are not provided. 

The experimental results are shown in Table~\ref{com_with_layoutllm} and Table~\ref{com_with_docllm}. Our method outperforms the baseline models, regardless of their type or whether the document's layout information is provided to them. It validates the rationale behind our design of multi-modal extension.

Additionally, our method significantly outperforms previous SOTA methods despite fewer training resources. As shown in Table~\ref{com_with_layoutllm}, it surpasses the performance of LayoutLLM~\cite{luo2024layoutllm} and LayTextLLM~\cite{lu2024laytextllm} though requiring fewer computational resources by using less data and tuning few parameters at training. Even with Llama2 as the backbone, our approach matches the performance of previous SOTA methods in the Document-oriented VQA task and significantly outperforms them in the KIE tasks, further demonstrating the efficiency and effectiveness of our approach.

Besides, our method exhibits robust zero-shot generalization capability without fine-tuning with specific downstream tasks. From the comparison with DocLLM~\cite{wang2024docllm} in Table~\ref{com_with_docllm}, it is clear that our method, even without fine-tuning, significantly outperforms DocLLM across most datasets and achieves a considerable overall performance improvement. This validates our design, which has preserved the strong generalization ability of the original LLM structure.

In summary, DocLayLLM serves as an efficient extension of LLMs. It demonstrates strong zero-shot generalization and effective document understanding by encoding documents within a unified model. Even with limited training resources, DocLayLLM achieves new SOTA performance compared to OCR-dependent competitors.

\subsection{Comparisons with SOTA OCR-free methods}
We have also compared our method with SOTA OCR-free TDU models. Table~\ref{com_with_textmonkey} presents the precision of our method and some advent OCR-free TDU methods across various benchmarks. Even though in the circumstance that OCR-free models use some of the training sets of these benchmarks during training, while our method infers in a purely zero-shot manner, our DocLayLLM still outperforms these models, particularly in KIE task, which heavily relies on layout information. This clearly demonstrates the superiority of our design to provide LLM with OCR information.

We additionally test DocLayLLM on a broader set of benchmarks from DocOwl 1.5~\cite{hu2024mplugdocowl1_5}. To further assess the performance of DocLayLLM on seen datasets, we incorporate the training sets of these benchmarks used in DocOwl 1.5 during the SFT process as other OCR-free TDU methods do during training. To maintain the efficiency advantage of our method, we ensure that the SFT is conducted with the same number of steps as the standard DocLayLLM. The results in Table~\ref{com_with_docowl} show that with the inclusion of downstream task data during SFT, our method outperforms the SOTA OCR-free TDU models in most datasets. Moreover, our overall performance in the Document-oriented VQA, KIE, and Table Understanding sub-tasks surpasses the OCR-free TDU methods using limited training resources, showcasing the tremendous potential of OCR-dependent methods and further reinforcing the reliability of our method.

Overall, our DocLayLLM consistently shows strong zero-shot inference ability, outperforming many OCR-free TDU methods without prior exposure to downstream data. Additionally, when incorporating downstream datasets during SFT like other OCR-free approaches, DocLayLLM surpasses them, highlighting its potential in solving TDU tasks by integrating OCR information into the LLM. 

\begin{figure}[!tbp]
\centering
\includegraphics[width=0.945\linewidth]{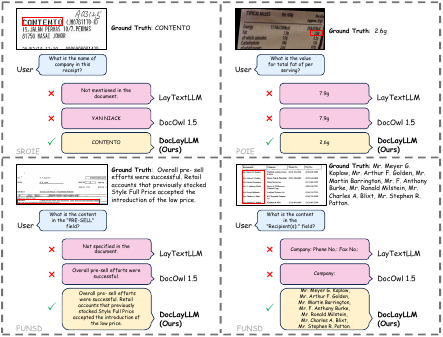}
\caption{Qualitative comparisons of DocLayLLM with other OCR-dependent and OCR-free methods. Zoom in for better view.}
\label{vis_1}
\end{figure}

\subsection{Qualitative Results}
Figure~\ref{vis_1} shows some qualitative examples to visualize the performance on KIE tasks of our DocLayLLM compared with LayTextLLM and DocOwl 1.5.
The example from SROIE clearly shows that DocLayLLM has improved document comprehension and answer accuracy. And the example from POIE demonstrates our advantage in handling table-formatted documents to give an accurate response. Furthermore, as seen in the FUNSD examples, DocLayLLM excels at extracting answers spanning multiple text lines, outperforming both OCR-dependent and OCR-free methods. For more visualization of qualitative examples, please refer to the appendix.

\subsection{Ablation Study}
We have also conducted additional experiments to explore the effectiveness of our proposed method. To conserve computational resources, we perform pre-training at 25\% of the standard scale (including the baseline) but full-scale SFT. We focus on the impact of two main factors: (a) the role of CoT pre-training and CoT annealing, and (b) the influence of different modality configurations.

Experimental results in Table~\ref{tab:ablation_setting} indicate that omitting CoT Pre-training leads to an obvious performance drop in both the Document-oriented VQA and KIE tasks, while not using CoT Annealing primarily affects VQA performance. Besides, using randomly initialized $VE$ and $LE$ partially compromises the proposed extension, resulting in performance degradation in both the VQA and KIE tasks.

Additionally, performances in Table~\ref{tab:ablation_modality} show the efficacy of our DocLayLLM to incorporate different modalities for TDU tasks. It is obvious that using only text from the document is insufficient for TDU, while adding layout information in textual form provides some gain on VQA tasks but leads to over-length input (see appendix for analysis), greatly slowing the inference. By our design to integrate layout features via an \textit{LE}, we achieve notable improvements on these TDU tasks with shorter inputs. And adding visual features further boosts the performance of our method.

Overall, the combined impact of these designs ensures the efficiency and effectiveness of our method.
\section{Conclusion}
\label{sec:conclusion}

\begin{table}[!tbp]
    \centering
    \begin{subtable}[t]{0.45\textwidth}
        \centering
        \resizebox{1\linewidth}{!}{
            \begin{tabular}{l|ccc|ccc}
            \toprule[2pt]
                  \multirow{2}{*}{\textbf{Experiment}}  & \multicolumn{3}{c|}{\textbf{Document-oriented VQA}} & \multicolumn{3}{c}{\textbf{KIE}} \\
                  & DocVQA & VisualMRC & \textbf{Avg.} & DeepForm & KLC   & \textbf{Avg.} \\
            \midrule
            \rowcolor[rgb]{ .741,  .843,  .933} \textbf{DocLayLLM$_{Llama3}$} & \textbf{77.43} & \textbf{316.27} & \textbf{196.85} & \textbf{35.41} & \textbf{26.25} & \textbf{30.83} \\
            \makecell[l]{\ \ \textit{w/o CoT Pre-training}} & 76.00  & 295.49  & 185.75  & 24.54  & 25.49  & 25.02  \\
            \makecell[l]{\ \ \textit{w/o CoT Annealing}} & 77.40  &  310.17  & 193.79   & 34.23  & 26.21   & 30.22   \\
            \makecell[l]{\ \ \textit{w/o pt VE \& LE}} & 77.42  & 308.49  & 192.96  & 31.31  & 25.76  & 28.54  \\
            \bottomrule[2pt]
            \end{tabular}
        }
        \caption{Impact of different settings: \textit{\textbf{w/o CoT Pre-training}} and \textit{\textbf{w/o CoT Annealing}} indicate pre-training without CoT and supervised fine-tuning with only CoT data, while \textit{\textbf{w/o pt VE and PE}} uses randomly initialized visual and layout embedders instead of pre-trained ones.}
        \label{tab:ablation_setting}
    \end{subtable}
    
    \bigskip
    
    \begin{subtable}[t]{0.45\textwidth}
        \centering
        \resizebox{1\linewidth}{!}{
            \begin{tabular}{ccc|ccc|ccc}
            \toprule[2pt]
            \multicolumn{3}{c|}{\textbf{Modality}} & \multicolumn{3}{c|}{\textbf{Document-oriented VQA}} & \multicolumn{3}{c}{\textbf{KIE}} \\
            \textbf{T }    &\textbf{ L}     &\textbf{ V }    & DocVQA & VisualMRC & \textbf{Avg.} & DeepForm & KLC   & \textbf{Avg.} \\
            \midrule
            \ding{51}     &  \ding{55}    &  \ding{55}    & 73.57  & 280.27  & 176.93  & 15.76  & 24.09  & 19.93  \\
            \ding{51}     & (\uppercase\expandafter{\romannumeral 1})     &  \ding{55}   & 75.72  & 306.51  & 191.12  & 14.13  & 24.23  & 19.18  \\
            \ding{51}     & (\uppercase\expandafter{\romannumeral 2})     &  \ding{55}    & 77.32  & 310.75  & 194.03  & 33.75  & 25.97  & 29.86  \\
            \rowcolor[rgb]{ .741,  .843,  .933} \ding{51}     & (\uppercase\expandafter{\romannumeral 2})     & \ding{51}    & \textbf{77.43} & \textbf{316.27} & \textbf{196.85} & \textbf{35.41} & \textbf{26.25} & \textbf{30.83} \\
            \bottomrule[2pt]
            \end{tabular}
        }
        \caption{Impacts of different modality configurations. \textbf{T}, \textbf{L}, and \textbf{V} represent textual, layout, and visual modalities, respectively. \textbf{(\uppercase\expandafter{\romannumeral 1})} indicates inputting OCR coordinates as text following \citet{he2023icld3ie}, while \textbf{(\uppercase\expandafter{\romannumeral 2})} embeds coordinates with $LE$ in our approach.}
        \label{tab:ablation_modality}
    \end{subtable}
    \caption{Ablation study on various designs of our DocLayLLM.}
    \label{tab:ablation}
\end{table}

We propose DocLayLLM, an efficient multi-modal extension of large language models for text-rich document understanding. By streamlined integration of document patch tokens and OCR positional tokens into the LLM inputs, it achieves a simple yet efficient multi-modal extension of LLM to enable it to have the capability to perceive and process textual, visual, and layout information from the OCR results of the document.
Additionally, we introduce CoT Pre-training and CoT Annealing techniques centered around CoT, offering a new way to generate CoT data and further enhancing the model's understanding of text-rich documents.
Despite requiring fewer training resources, our method surpasses existing OCR-dependent TDU models, demonstrating strong zero-shot generalization capability. Furthermore, it also exceeds the performance of SOTA OCR-free TDU methods under comparable training conditions, showcasing remarkable performance across various document understanding tasks. 
Our work significantly advances efficient MLLM-based document understanding while maintaining high-performance standards.
\section*{Acknowledgement}
This research is supported in part by National Natural Science Foundation of China (Grant No.: 62476093, 62441604).
{
    \small

    \bibliographystyle{ieeenat_fullname}
   
    \bibliography{main}
}
\clearpage
\maketitlesupplementary
\setcounter{table}{0}
\setcounter{figure}{0}
\setcounter{section}{0}
\renewcommand{\thetable}{\Alph{table}}
\renewcommand{\thefigure}{\Alph{figure}}
\renewcommand{\thesection}{\Alph{section}}

\section{Pre-traing tasks}
\label{sec:pre-training_tasks}
We adopt pre-training tasks as shown in Table~\ref{pretraining_task}. These tasks facilitate the alignment of layout and visual features with the LLM’s feature space while enhancing the LLM’s understanding of document content.

\begin{table}[h]
\centering
\resizebox{0.95\linewidth}{!}{
    \begin{tabular}{>{\RaggedRight\hangindent=1em}p{0.5\linewidth} >{\RaggedRight\hangindent=1em}p{0.8\linewidth}}
    \toprule[2pt]
    \textbf{Task} & \textbf{Description} \\
    \midrule
    Document Description & Provide a brief overview of the document. \\
    Text and Box Reconstruction & Recover the coordinates of bounding boxes of all the OCR text. \\
    Layout Analysis & Determine the layout type (e.g., Title, Author, Paragraph, etc) of a giving area or locate specific layout elements. \\
    Table Analysis & Decode the structure of tables and identify the positions of elements within. \\
    Mask Language Model & Restore masked words in the provided OCR text. \\
    Mask Position Model & Reconstruct the box for text elements missing the coordinates of the bounding box. \\
    Geometric Analysis & Calculate distances or directions between two specified text elements. \\
    \bottomrule[2pt]
    \end{tabular}
}
\caption{The training tasks during the pre-training stage.}
\label{pretraining_task}
\end{table}

\section{More Implementation Details}
\label{sec:implement_details}

We implement our DocLayLLM using Llama2-7B-Chat~\citep{touvron2023llama2} and Llama3-8B-Instruct~\citep{llama32024llama3_1}. The hyper-parameters for both pre-training and supervised fine-tuning are detailed in Table~\ref{hyper_params}. As shown in the table, our DocLayLLM demonstrates efficiency, requiring fewer training resources while maintaining high performance. This underscores the method's capability to deliver robust results without the need for extensive computational power, making it a resource-efficient solution for text-rich document understanding tasks.

\begin{table}[ht]
\centering
\resizebox{0.85\linewidth}{!}{
    \begin{tabular}{lcc}
    \toprule[2pt]
        \textbf{Parameters}  & \textbf{Pre-Training} & \textbf{Supervised Fine-Tuning}  \\
    \midrule
    \textbf{LoRA Rank} & 64  & 64 \\
    \textbf{Batch Size} & 512  & 64 \\
    \textbf{Max Length} & 2560  & 2560 \\
    \textbf{Precision} & bf16  & bf16 \\
    \makecell[l]{\textbf{Trainable}\\ \textbf{\ \ \ \ Parameters}}  & \makecell[c]{170M/Llama2;\\ 178M/Llama3}  & \makecell[c]{170M/Llama2;\\ 178M/Llama3} \\
    \makecell[l]{\textbf{Fixed}\\ \textbf{\ \ \ \ Parameters}} & \makecell[c]{6.7B/Llama2;\\ 8.0B/Llama3}   & \makecell[c]{6.7B/Llama2;\\ 8.0B/Llama3}  \\
    \textbf{Learning Rate} & 1e-4  & 2e-5 \\
    \textbf{Weight Decay} & 0.01  & 0.01 \\
    \textbf{Scheduler} & cosine  & cosine \\
    \textbf{Adam Betas} & [0.9, 0.999]  & [0.9, 0.999] \\
    \textbf{Adam Epsilon} & 1e-8  & 1e-8 \\
    \textbf{Epoch} & 1  & 3 \\
    \bottomrule[2pt]
    \end{tabular}
}
\caption{Hyper-parameters of DocLayLLM.}
\label{hyper_params}
\end{table}

\section{More Qualitative Examples}
\label{more_qualitative_examples}

\begin{figure*}[htbp]
    \centering
    \begin{subfigure}[b]{\linewidth}
        \centering
        \includegraphics[width=0.95\linewidth]{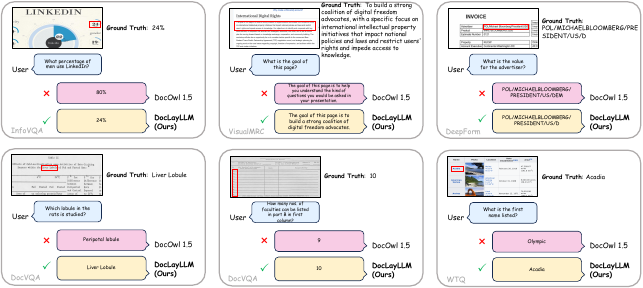}
        \caption{Qualitative comparisons with DocOwl 1.5~\citep{hu2024mplugdocowl1_5} across various benchmarks. The document-oriented VQA tasks include InfoVQA~\citep{mathew2021infovqa}, VisualMRC~\citep{tanaka2021visualmrc}, and DocVQA~\citep{mathew2021docvqa}; the KIE task includes DeepForm~\citep{svetlichnaya2020deepform}; and the Table Understanding task includes WTQ~\citep{pasupat2015wtq}.}
        \label{fig:comp_with_docowl1_5}
    \end{subfigure}
    
    \bigskip
    
    \begin{subfigure}[b]{\linewidth}
        \centering
        \includegraphics[width=0.95\linewidth]{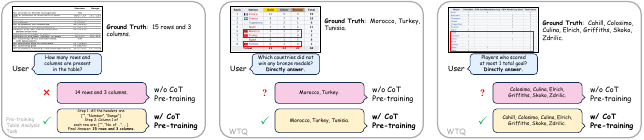}
        \caption{Qualitative comparisons between the use and absence of CoT Pre-training. \textbf{w/o Pre-training} indicates the absence of CoT at the pre-training stage, while \textbf{w/ CoT Pre-training} denotes its application. \textbf{``\textcolor{red}{?}"} represents that the answer is ambiguous.}
        \label{fig:comp_of_cot_pretraining}
    \end{subfigure}
    
    \bigskip
    
    \begin{subfigure}[b]{\linewidth}
        \centering
        \includegraphics[width=0.95\linewidth]{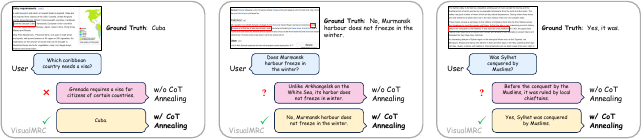}
        \caption{Qualitative comparisons between the use and absence of CoT Annealing. \textbf{w/o CoT Annealing} indicates the absence of CoT Annealing, while \textbf{w/ CoT Annealing} denotes its application. \textbf{``\textcolor{red}{?}"} represents that the answer is ambiguous.}
        \label{fig:comp_of_cot_annealing}
    \end{subfigure}
    \caption{Further qualitative comparisons of DocLayLLM against the SOTA OCR-free method and under various settings.}
    \label{vis_2}
\end{figure*}

We also provide additional qualitative examples of our DocLayLLM. As shown in the comparison between DocLayLLM and the SOTA OCR-free method DocOwl 1.5~\citep{hu2024mplugdocowl1_5} in Figure~\ref{fig:comp_with_docowl1_5}, our DocLayLLM demonstrates superior document understanding capabilities, delivering accurate answers in examples from InfoVQA and VisualMRC. Furthermore, in the DeepForm example, we observe that our design to integrate OCR information helps reduce the occurrence of hallucinated outputs compared to OCR-free methods. Moreover, in the examples from DocVQA and WTQ, DocLayLLM reliably exhibits robust table comprehension abilities. These results collectively highlight the effectiveness of our design in incorporating OCR information.

Furthermore, we also present the results of whether Table-Structure-Aware CoT is used during the pre-training stage. As shown in Figure~\ref{fig:comp_of_cot_pretraining}, models incorporating CoT demonstrate a more comprehensive understanding of table structures, leading to more accurate and thorough answers in table-related downstream document understanding tasks. This validates the effectiveness of our proposed CoT pre-training approach.

Additionally, we visualized the outputs with and without the use of CoT Annealing. As shown in the visualization of VisualMRC in Figure~\ref{fig:comp_of_cot_annealing}, DocLayLLM employing CoT Annealing tends to provide more straightforward and accurate answers. This is particularly evident in yes-or-no questions, where the model without CoT Annealing often fails to directly respond with a clear "yes" or "no" but repeats the sentence in the document where the answer is located. In contrast, the model using CoT Annealing typically provides a direct answer first, followed by an explanation. These observations indicate that CoT Annealing enables the model to answer questions more directly, thereby enhancing its performance.

\newpage
\section{Input Length Analysis}
\label{input_length_analysis}

In our ablation study, we evaluated the performance of different methods for incorporating OCR information. This section further examines the input length of OCR information under various approaches. The analysis was conducted using Llama3~\citep{llama32024llama3_1} as the base model, with its tokenizer applied for tokenization. Table~\ref{tab:len_analysis} presents a comparison of the average input length of OCR information across several benchmarks under two configurations: \textbf{(\uppercase\expandafter{\romannumeral 1})} encoding OCR bounding box coordinates as plain text, following the approach of ICL-D3IE~\citep{he2023icld3ie}, and \textbf{(\uppercase\expandafter{\romannumeral 2})} encoding OCR bounding box coordinates using a layout embedder $LE$.

The results clearly show that encoding with $LE$ significantly reduces the input length, thereby enhancing efficiency during both training and inference. These findings underscore the efficiency of our proposed DocLayLLM.

\begin{table}[ht]
\centering
\resizebox{0.9\linewidth}{!}{
    \begin{tabular}{c|cc|cc}
    \toprule[2pt]
    \multirow{2}{*}{\makecell[c]{\textbf{Input}\\ \textbf{Method}}} & \multicolumn{2}{c|}{\textbf{Document-oriented VQA}} & \multicolumn{2}{c}{\textbf{KIE}} \\
        & DocVQA & VisualMRC & DeepForm & KLC  \\
    \midrule
    \textbf{(\uppercase\expandafter{\romannumeral 1})} & 1571.80  & 6269.17  & 4952.87  & 457.58  \\
    \rowcolor[rgb]{ .741,  .843,  .933} \textbf{(\uppercase\expandafter{\romannumeral 2})} & \textbf{455.17}  & \textbf{2095.35}  & \textbf{1198.97}  & \textbf{125.40}  \\
    \bottomrule[2pt]
    \end{tabular}
}
\caption{The average input length of OCR information across various benchmarks, comparing different ways to input OCR bounding box coordinates.}
\label{tab:len_analysis}
\end{table}

\section{OCR Result Impacts}
Since DocLayLLM requires OCR result input, we explored the impact of OCR quality on the performance of DocLayLLM. In the results presented in the main text, we used the official OCR results when evaluating on the DocVQA benchmark. To assess the model's applicability in real-world scenarios, we employed a commercial OCR engine~\footnote{\href{https://www.textin.com/}{https://www.textin.com/}} to process DocVQA and used the recognized text for further testing DocLayLLM’s performance. The results in Table~\ref{tab:ocr error} suggest that the reported results in the main text have not fully reflected the potential of DocLayLLM. The model could achieve even better performance with real-world OCR results.

\begin{figure}[ht]
\centering
\includegraphics[width=0.945\linewidth]{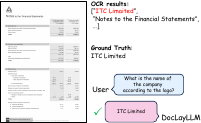}
\caption{
Illustration of DocLayLLM's OCR error correction capability.
}
\label{fig:ocr_error}
\end{figure}

Furthermore, as illustrated in Figure~\ref{fig:ocr_error}, we observed that when OCR errors occur, DocLayLLM has the capability to correct these errors and produce the final correct answer. This further substantiates the robustness of DocLayLLM in real-world scenarios.

\newpage
\begin{table}[t]
\centering
\resizebox{0.45\linewidth}{!}{
    \begin{tabular}{l|c}
    \toprule[2pt]
          &  ANLS$\uparrow$  \\
    \midrule
    Official OCR & 86.52 \\
    Commercial Engine & \textbf{87.52} \\
    \bottomrule[2pt]
    \end{tabular}
}
\caption{Performances of DocLayLLM on DocVQA benchmark with different OCR results.}
\label{tab:ocr error}
\end{table}
\ 

\end{document}